\def\cvprPaperID{5707}
\def\confName{ICCV}
\def\confYear{2023}

\def\paperTitle{Resolving Multi-Condition Confusion for Finetuning-Free Personalized Image Generation}

\def\authorBlock{
    Qihan Huang\textsuperscript{$1,2,*$},
    Siming Fu\textsuperscript{$2,*$},
    Jinlong Liu\textsuperscript{$2$},
    Hao Jiang\textsuperscript{$2$},
    Yipeng Yu\textsuperscript{$2$},
    Jie Song\textsuperscript{$1,\dagger$} \\
    \textsuperscript{$1$}Zhejiang University,
    \textsuperscript{$2$}Alibaba Group \\
    
    {\tt\small \{qh.huang,sjie\}@zju.edu.cn}, \\
    {\tt\small fusiming.fsm@taobao.com}, {\tt\small LJLwykqh@126.com}, {\tt\small aoshu.jh@alibaba-inc.com}, {\tt\small yypzju@163.com}
}

\newif\ifreview 
\newif\ifarxiv 
\newif\ifcamera \newcommand{\cameraready}{\cameratrue}
\newif\ifrebuttal

\newcommand{\mybold}[1]{\boldsymbol{#1}}
\cameraready

\pdfoutput=1
\documentclass[10pt,twocolumn,letterpaper]{article}
\ifreview \usepackage[review]{cvpr} \fi
\ifarxiv \usepackage[pagenumbers]{cvpr} \fi
\ifrebuttal \usepackage[rebuttal]{cvpr} \fi
\ifcamera \usepackage{cvpr} \fi

\usepackage{graphicx}
\usepackage{amsmath}
\usepackage{amssymb}
\usepackage{booktabs}

\usepackage{times}
\usepackage{microtype}
\usepackage{epsfig}
\usepackage[table,xcdraw]{xcolor}
\usepackage{caption}
\usepackage{float}
\usepackage{placeins}
\usepackage{color, colortbl}
\usepackage{stfloats}
\usepackage{enumitem}
\usepackage{tabularx}
\usepackage{xstring}
\usepackage{multirow}
\usepackage{xspace}
\usepackage{url}
\usepackage{subcaption}
\usepackage{xcolor}
\usepackage[hang,flushmargin]{footmisc}
\usepackage{bbm}
\usepackage{graphicx}
\usepackage{amsthm}
\usepackage{makecell}
\usepackage{bbding}
\usepackage{amsmath}
\definecolor{mygreen}{RGB}{52, 157, 2}

\usepackage{paralist}
\usepackage[accsupp]{axessibility}
\ifcamera \usepackage[accsupp]{axessibility} \fi

% ...

\ifarxiv  \fi

\newcommand{\R}[1]{{%
    \textbf{%
        \ifstrequal{#1}{1}{\textcolor{red}{R#1}}{%
        \ifstrequal{#1}{2}{\textcolor{blue}{R#1}}{%
        \ifstrequal{#1}{3}{\textcolor{magenta}{R#1}}{%
        \ifstrequal{#1}{4}{\textcolor{teal}{R#1}}{%
                           \textcolor{cyan}{R#1}%
        }}}}%
    }%
}}

\usepackage{xr-hyper}

\makeatletter
\newcommand*{\addFileDependency}[1]{
  \typeout{(#1)}
  \@addtofilelist{#1}
  \IfFileExists{#1}{}{\typeout{No file #1.}}
}

\makeatother

\usepackage[pagebackref,breaklinks,colorlinks]{hyperref}
\usepackage[capitalize]{cleveref}
\crefname{section}{Sec.}{Secs.}
\crefname{table}{Tab.}{Tabs.}
\crefname{figure}{Fig.}{Figs.}

\begin{document}
\title{\paperTitle}
\author{\authorBlock}
\maketitle
\footnotetext[1]{$*$ Equal contribution.}
\footnotetext[2]{$\dagger$ Corresponding author.}

\begin{abstract}
Personalized text-to-image generation methods can generate customized images based on the reference images, which have garnered wide research interest.
Recent methods propose a finetuning-free approach with a decoupled cross-attention mechanism to generate personalized images requiring no test-time finetuning.
However, when multiple reference images are provided, the current decoupled cross-attention mechanism encounters the \textit{object confusion} problem and fails to map each reference image to its corresponding object, thereby seriously limiting its scope of application.
To address the object confusion problem, in this work we investigate the relevance of different positions of the latent image features to the target object in diffusion model, and accordingly propose a weighted-merge method to merge multiple reference image features into the corresponding objects.
Next, we integrate this weighted-merge method into existing pre-trained models and continue to train the model on a multi-object dataset constructed from the open-sourced SA-1B dataset.
To mitigate object confusion and reduce training costs, we propose an \textit{object quality score} to estimate the image quality for the selection of high-quality training samples.
Furthermore, our weighted-merge training framework can be employed on single-object generation when a single object has multiple reference images.
The experiments verify that our method achieves superior performance to the state-of-the-arts on multi-object personalized image generation, and remarkably improves the performance on single-object personalized image generation.
Our code is available at \textit{~\url{https://github.com/hqhQAQ/MIP-Adapter}}.
\end{abstract}

\section{Introduction}
\label{sec:intro}

Personalized text-to-image generation methods generate images conditioned on the reference images that specify the details of the generated contents, sparking considerable research interest due to its diverse applications.
The methodology in this domain is gradually shifting from a \textit{finetuning-based} approach~(\textit{e.g.}, DreamBooth~\cite{ruiz2023dreambooth}, Custom Diffusion~\cite{kumari2023custom_diffusion}) to a \textit{finetuning-free} technique~(\textit{e.g.}, IP-Adapter~\cite{ye2023ip}, Subject-Diffusion~\cite{ma2024subject_diffusion}), as finetuning-free methods eliminate the need for finetuning during test time and significantly reduce the usage cost.
%


\begin{figure}[t]
\centering
    \includegraphics[width=\linewidth]{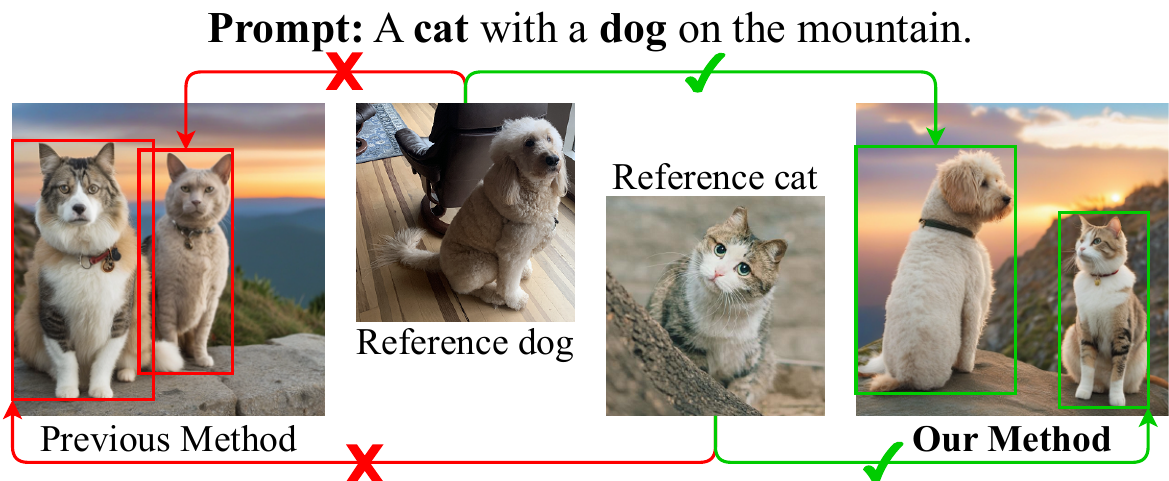}
\caption{Left image demonstrates the object confusion problem in decoupled cross-attention mechanism, and right image presents the correct generation using our method.}
\vspace{-1em}
\label{fig:intro1}
\end{figure}

\begin{figure}[t]
\centering
    \includegraphics[width=\linewidth]{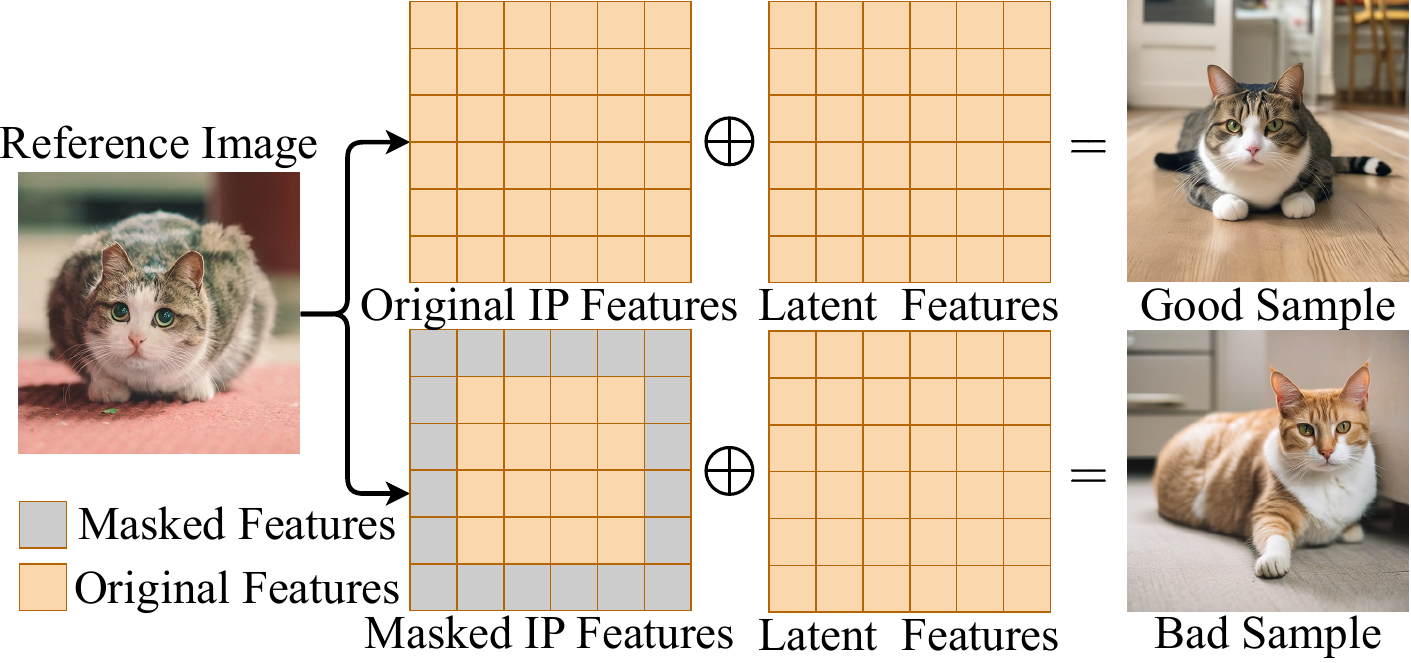}
\caption{The reference image~(IP) features, with the background masked, reduces generation quality in IP-Adapter.}
\label{fig:intro2}
\vspace{-1em}
\end{figure}

Early finetuning-free methods, such as InstantBooth~\cite{shi2024instantbooth} and FastComposer~\cite{xiao2023fastcomposer}, simply integrate the features of the reference image into the text embeddings and feed them into the text encoder, without fully exploiting the information from the reference image.
Recent finetuning-free methods, such as IP-Adapter~\cite{ye2023ip}, more comprehensively utilize the features of the reference image by training additional cross-attention layers to integrate reference image features into the intermediate layers of the diffusion model, and achieve comparable performance to the finetuning-based methods.
However, the current decoupled cross-attention only considers one reference image for each generation. When multiple reference images are provided, the decoupled cross-attention suffers from the \textit{object confusion} problem if applied straightforwardly, wherein object features in the reference images are assigned to the wrong objects in the generated images, as illustrated in \autoref{fig:intro1}.
Some previous image generation methods~\cite{yang2024mastering} attempt to mitigate the object confusion issue by incorporating the object features into the corresponding regions of latent image features in the diffusion model.  
Nevertheless, as the object information is distributed over the entire image feature space rather than confined to the corresponding local region owing to large receptive fields in deep networks~\cite{luo2016understanding, araujo2019computing},
the generated images can be limited in faithfulness to the reference images~(\textit{i.e.}, the appearance differs between the generated and the reference images), as shown in \autoref{fig:intro2}.
In this work, rather than splitting latent features into different regions, we propose a \textbf{weighted-merge method} to merge the reference image features into the whole latent image features with different weights on different positions.
Specifically, this work estimates these weights as the relevance of different positions in latent image features to the target object, by ingeniously utilizing the cross-attention weights between the text features of the target object and the latent image features within the stable diffusion model.
Besides, we design an experiment that adds different noise to the latent image features based on the predicted object relevance, verifying the effectiveness of this object relevance estimation method.
We employ this method on the pre-trained finetuning-free personalized generation models~(\textit{e.g.}, IP-Adapter), enabling multi-object generation by simultaneously merging multiple conditions~(reference images \& text prompts) into the model. Experiment results indicate that our method can alleviate object confusion and significantly improve the performance of multi-object personalized image generation for these models without any training.

Although weighted-merge effectively alleviates object confusion, adding multiple reference images at once will interfere with the latent image features, causing them to deviate from their distribution in the original model and resulting in lower generation quality. To address this issue, this work trains the pre-trained finetuning-free model with the weighted-merge method on a multi-object dataset. Specifically, this dataset is constructed from the open-sourced SA-1B dataset~\cite{kirillov2023sam} consisting of about 11 million images with multiple objects.
Besides, this work proposes an object quality score to estimate the object quality of the image, according to the the degree of confusion between multiple objects, as well as the matching degree between object texts and images.
Based on the object quality score, we can select high-quality images that alleviate the object confusion problem for higher performance while decreasing training costs.

Moreover, this weighted-merge training framework can be applied to single-object generation, because a single object has multiple reference images in reality.  
Compared to previous approaches that only use a single reference image or simply average the features of multiple images, our weighted-merge method can extract diverse useful information from different reference images and adaptively merge them to achieve superior results.

We perform comprehensive experiments to validate the performance of our proposed framework.
Experiment results demonstrate that with only 100,000 high-quality images~(0.13\% of the dataset from Subject Diffusion) selected from SA-1B, our model achieves state-of-the-art performance on the Concept101 dataset and DreamBooth dataset of multi-object personalized image generation.
Besides, our weighted-merge training framework significantly improves the performance of pre-trained model on the DreamBooth dataset of single-object personalized image generation.

To sum up, the main contributions of this work can be summarized as follows:

$\bullet$ We extend the decoupled cross-attention mechanism of finetuning-free personalized image generation methods to merge multiple conditions, with a proposed weighted-merge method to tackle the object confusion problem.

$\bullet$ We construct a small but high-quality dataset from the open-sourced SA-1B dataset for model training, with a proposed object quality score for image selection.

$\bullet$ Experiment results demonstrate that our weighted-merge training framework outshines in merging multiple conditions, and our model achieves state-of-the-art performance on both the Concept101 dataset and DreamBooth dataset of multi-object personalized image generation.

\section{Related Work}

\noindent \textbf{Finetuning-Based Personalized Image Generation.}
Early personalized image generation methods require finetuning the original diffusion model on the reference images.
Specifically, DreamBooth finetunes the entire UNet network of diffusion model, Textual Inversion~\cite{gal2023textual_inversion} finetunes only the special embedding vector of the target object, and Custom Diffusion finetunes only the K and V layers of the cross-attention in the UNet network.
Cones~\cite{liu2023cones} detects the concept neurons in the K and V layers and updates them during training.
Mix-of-Show~\cite{gu2023mix_of_show} trains a separate LoRA model for each object and merges them with gradient fusion.
However, these methods require finetuning for each object, which consumes a lot of computational resources and is not suitable for real applications.

\noindent \textbf{Finetuning-Free Personalized Image Generation.}
Finetuning-free methods train the model to directly incorporate the reference image features on a large dataset, without the need for additional finetuning during test time.
Early finetuning-free methods~(\textit{e.g.}, InstantBooth, FastComposer, and Taming Encoder~\cite{jia2023taming}) simply integrate the image features into the text embeddings, without fully utilizing the reference image information.
Recent methods~(\textit{e.g.}, IP-Adapter, ELITE~\cite{wei2023elite}, and SSR-Encoder~\cite{zhang2024ssr}) make more extensive utilization of reference image information by integrating the image features into the middle layers of the diffusion model, using a decoupled cross-attention mechanism.
These methods excel at merging a single reference image and achieve impressive performance.
However, decoupled cross-attention encounters the object confusion problem when merging multiple reference images, a problem this study aims to address.

\begin{figure*}[t]
\centering
    \includegraphics[width=\linewidth]{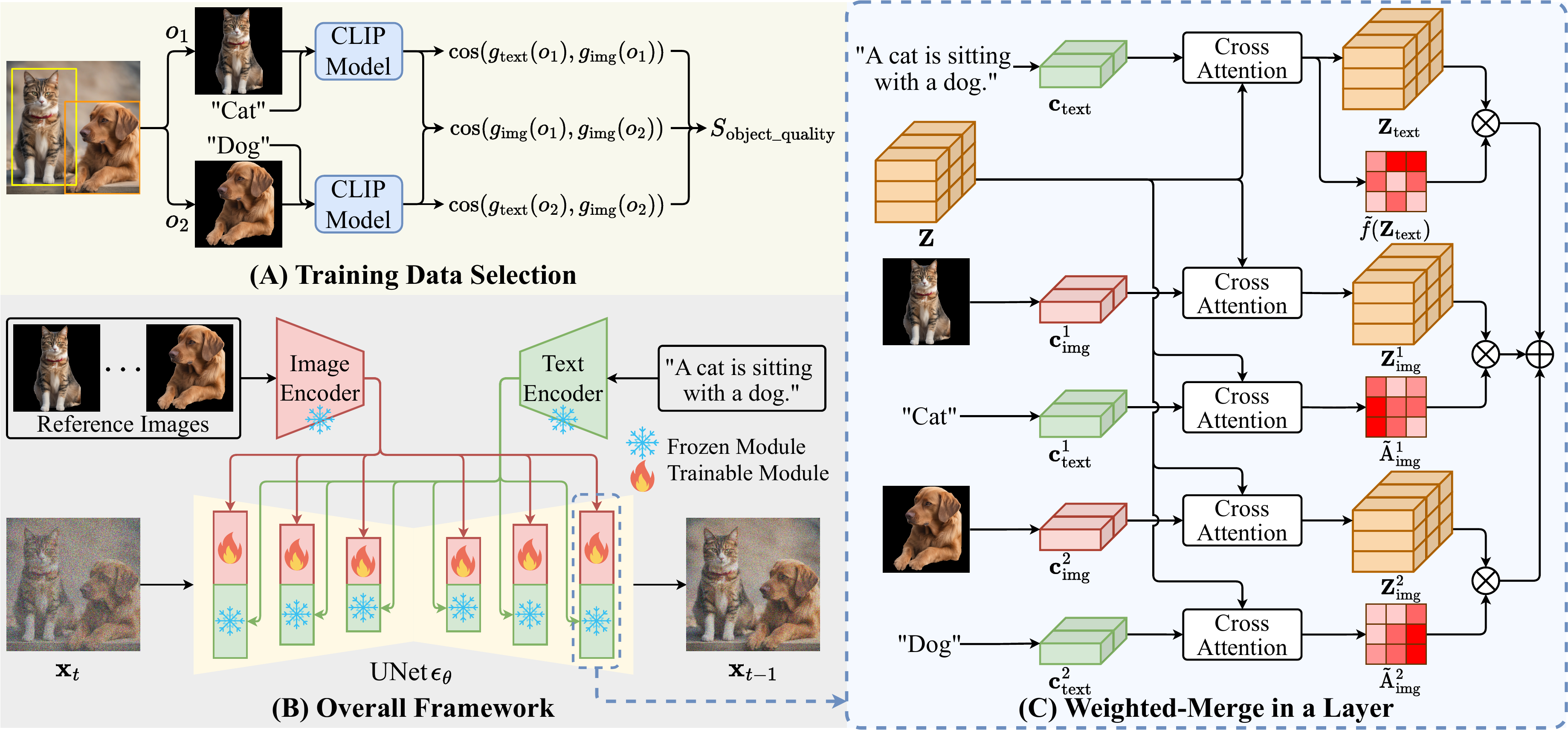}
\caption{
\textbf{(A)} demonstrates the calculation of $S_{\rm object\_relevance}$, which is used for selecting training data.
The overall framework in \textbf{(B)} consists of a UNet model for noise prediction conditioned on the text prompt and multiple reference images.
\textbf{(C)} presents the proposed weighted-merge method in each cross-attention layer of UNet from \textbf{(B)}.
$\tilde {\rm A}_{\rm img}^{i} = \frac{ {\rm A}_{\rm img}^{i} }{ \bar {\rm A}_{\rm img}^{i} }$, and $\tilde f(\mathbf{Z}_{\rm text}) = \frac{ f(\mathbf{Z}_{\rm text}) }{ \bar f(\mathbf{Z}_{\rm text}) }$.
}
\vspace{-1em}
\label{fig:method}
\end{figure*}

\section{Method}

In this section, we first give the preliminaries in \textbf{section 1}, then propose the object relevance estimation method in \textbf{section 2}.
Next, \textbf{section 3} and \textbf{section 4} propose the weighted-merge method and directly apply it to the current pre-trained model.
Finally, \textbf{section 5} proposes the training framework for further performance improvement.

\subsection{1. Preliminaries \label{sec:preliminaries}}

\textbf{Diffusion model.}
Current personalized image generation methods adopt diffusion model~\cite{ho2020ddpm, rombach2022ldm} as the base model.
Diffusion model consists of two processes: a diffusion process which gradually adds noise into the original image with a Markov chain in $T$ steps, and a denoising process which predicts the noise to generate the image using a deep neural network.
Specifically, personalized image generation methods generate images simultaneously conditioned on the text prompt and the reference images.
Typically, $\mybold{\epsilon}_{\theta}$ denotes the deep neural network for noise prediction, and the training loss of personalized diffusion model is defined as below:

\begin{equation}
    \mathcal{L} = \mathbb{E}_{\mybold{x}_0, \mybold{\epsilon} \in \mathcal{N}(\mathbf{0}, \mathbf{I}), \mybold{c}_{\rm text}, \mybold{c}_{\rm img}} \| \mybold{\epsilon} - \mybold{\epsilon}_{\theta}(\mybold{x}_t, \mybold{c}_{\rm text}, \mybold{c}_{\rm img}, t) \|^{2},
\nonumber
\end{equation}

where $\mybold{x}_0$ denotes the original real image, $t \in [0, T]$ denotes the time step in the diffusion process,
$\mybold{x}_t = \alpha_t \mybold{x}_0 + \sigma_t \mybold{\epsilon}$, and $\alpha_t$, $\sigma_t$ are predefined weights for step $t$ in the diffusion process.
$\mybold{c}_{\rm text}$ denotes the text features, and $\mybold{c}_{\rm img}$ denotes the reference image features.
After training, the model can generate images by progressively denoising Gaussian noise in multiple steps.

\vspace{0.5em}
\noindent \textbf{Decoupled cross-attention mechanism.}
Recent finetuning-free personalized image generation methods adopt decoupled cross-attention to merge the text features and reference image features into the middle layers of model $\mybold{\epsilon}_{\theta}$.
Specifically, the latent image features $\mathbf{Z} \in \mathbb{R}^{(H \cdot W) \times D}$ in a middle layer are fed into a cross-attention module to interact with the text features $\mybold{c}_{\rm text} \in \mathbb{R}^{S_{\rm text} \times D_{\rm text}}$:

\begin{equation}
    \mathbf{Z}_{\rm text} \!=\! \mathrm{Attn}(\mathbf{Q}, \mathbf{K}_{\rm text}, \mathbf{V}_{\rm text}) \!=\! \mathrm{Softmax}(\frac{\mathbf{Q} \mathbf{K}_{\rm text}^{\top}}{\sqrt{d}}) \mathbf{V}_{\rm text}.
    \nonumber
\end{equation}

Here, $\mathbf{Q} = \mathbf{Z} \mathbf{W}^{\mathbf{Q}}$, $\mathbf{K}_{\rm text} = \mybold{c}_{\rm text} \mathbf{W}_{\rm text}^{\mathbf{K}}$, $\mathbf{V}_{\rm text} = \mybold{c}_{\rm text} \mathbf{W}_{\rm text}^{\mathbf{V}}$ are the query, key, and value matrices of the attention operation, respectively, and $\mathbf{W}^{\mathbf{Q}} \in \mathbb{R}^{D \times D}$, $\mathbf{W}_{\rm text}^{\mathbf{K}} \in \mathbb{R}^{D_{\rm text} \times D}$, $\mathbf{W}_{\rm text}^{\mathbf{V}} \in \mathbb{R}^{D_{\rm text} \times D}$ are the learnable weight matrices for feature projection.
Besides, $\mathbf{Z}$ is also fed into another cross-attention module to interact with the reference image features $\mybold{c}_{\rm img} \in \mathbb{R}^{S_{\rm img} \times D_{\rm img}}$:

\begin{equation}
    \mathbf{Z}_{\rm img} \!=\! \mathrm{Attn}(\mathbf{Q}, \mathbf{K}_{\rm img}, \mathbf{V}_{\rm img}) \!=\! \mathrm{Softmax}(\frac{\mathbf{Q} \mathbf{K}_{\rm img}^{\top}}{\sqrt{d}}) \mathbf{V}_{\rm img}.
    \nonumber
\end{equation}

Likewise, $\mathbf{K}_{\rm img} = \mybold{c}_{\rm img} \mathbf{W}_{\rm img}^{\mathbf{K}}$, $\mathbf{V}_{\rm img} = \mybold{c}_{\rm img} \mathbf{W}_{\rm img}^{\mathbf{V}}$, and $\mathbf{W}_{\rm img}^{\mathbf{K}} \in \mathbb{R}^{D_{\rm img} \times D}$, $\mathbf{W}_{\rm img}^{\mathbf{V}} \in \mathbb{R}^{D_{\rm img} \times D}$ are the learnable weight matrices for projecting the reference image features.
Next, the final output of the decoupled cross-attention $\mathbf{Z}_{\rm new}$ is the addition of $\mathbf{Z}_{\rm text}$ and $\mathbf{Z}_{\rm img}$:

\begin{equation}
    \mathbf{Z}_{\rm new} = \mathbf{Z}_{\rm text} + \mathbf{Z}_{\rm img}.
    \nonumber
\end{equation}

\subsection{2. Object Relevance Estimation \label{sec:object_relevance_estimation}}

Decoupled cross-attention mechanism excels at merging a single reference image into the model $\mybold{\epsilon}_{\theta}$ than previous methods that only add the reference image features into the text embeddings.
However, decoupled cross-attention simply merges the text-conditioned latent image features $\mathbf{Z}_{\rm text}$ and image-conditioned latent image features $\mathbf{Z}_{\rm img}$ with an addition operation, without constraining the reference image to the corresponding object in the text prompt.
This results in an \textit{object confusion problem} when merging multiple reference images, which incorrectly adds the reference image information to its unrelated objects, as shown in \autoref{fig:intro1}.
Therefore, given $M$ reference images corresponding to $M$ objects in the text prompt, this work strives to merge $M$ image-conditioned latent image features $\{ \mathbf{Z}_{\rm img}^{i} \}_{i=1}^{M}$ into the text-conditioned latent image features $\mathbf{Z}_{\rm text}$ by resolving the object confusion problem.

To this end, this work first investigates the information distribution of an object~(as referenced in the text prompt) on $\mathbf{Z}_{\rm text}$.
Some methods~(\textit{e.g.}, RPG~\cite{yang2024mastering}) assume that the position of the object in $\mathbf{Z}_{\rm text}$ is the same as that in the generated image, however, this assumption is not accurate.
Actually, the deep neurons in the deep neural networks have large effective receptive fields~\cite{luo2016understanding, araujo2019computing}, meaning that a wide range of latent image features can affect the target object in the generated image, rather than being limited to only the local latent image features with the same position as the target object.
As shown in \autoref{fig:intro2}, $\mathbf{Z}_{\rm img}$, with the background masked, will decrease the generation quality of the foreground cat in the generated image.
Therefore, simply adding the reference image information into some local regions of $\mathbf{Z}_{\rm text}$ will lead to information loss and degrade the performance.

To tackle this problem, we estimate the relevance of \textbf{all positions} in $\mathbf{Z}_{\rm text}$ to the target object, and merge $\mathbf{Z}_{\rm img}^{i}$ into each position of $\mathbf{Z}_{\rm text}$ with different weights according to the estimated relevance.
For estimating the object relevance, this work ingeniously utilizes the original cross-attention modules within model $\mybold{\epsilon}_{\theta}$ and calculates the attention map between the text features of the object and the original latent image features $\mathbf{Z}$~(note that $\mathbf{Z}_{\rm text}$ is calculated from $\mathbf{Z}$).
Specifically, we first extract the text features $\mybold{c}_{\rm text}^{i} \in \mathbb{R}^{S_{\rm text} \times D_{\rm text}}$ of the $i$-th object~(corresponding to the $i$-th reference image) by feeding the object text into the text encoder.
Next, the object relevance ${\rm A}_{\rm img}^{i} \in \mathbb{R}^{(H \cdot W)}$ of the $i$-th object to $\mathbf{Z}_{\rm text} \in \mathbb{R}^{(H \cdot W) \times D}$ is calculated by averaging the original cross-attention matrix:

\begin{equation}
    {\rm A}_{\rm img}^{i} = \frac{1}{S_{\rm text}} \sum\limits_{j=1}^{S_{\rm text}} \mathrm{Softmax}(\frac{\mathbf{K}_{\rm text}^{i} \mathbf{Q}^{\top}}{\sqrt{d}})[j],
    \nonumber
\end{equation}

where $\mathbf{Q} = \mathbf{Z} \mathbf{W}^{\mathbf{Q}}$, $\mathbf{K}_{\rm text}^{i} = \mybold{c}_{\rm text}^{i} \mathbf{W}_{\rm text}^{\mathbf{K}}$~(note that $\mathbf{W}_{\rm text}^{\mathbf{K}}$ is shared with the original text features $\mybold{c}_{\rm text}$), and $\mathrm{Softmax}(\frac{\mathbf{K}_{\rm text}^{i} \mathbf{Q}^{\top}}{\sqrt{d}})[j] \in \mathbb{R}^{(H \cdot W)}$ is the $j$-th element of $\mathrm{Softmax}(\frac{\mathbf{K}_{\rm text}^{i} \mathbf{Q}^{\top}}{\sqrt{d}}) \in \mathbb{R}^{S_{\rm text} \times (H \cdot W)}$.

\subsection{3. Training-Free Personalized Image Generation \label{sec:weighted_merge}}

Based on the above object relevance estimation method, we propose a weighted-merge method to extend current pre-trained models~(\textit{e.g.}, IP-Adapter) to multi-object personalized image generation, in a training-free manner.   
Specifically, this method first generates the text-conditioned latent image features $\mathbf{Z}_{\rm text}$ and $M$ image-conditioned latent image features $\{ \mathbf{Z}_{\rm img}^{i} \in \mathbb{R}^{(H \cdot W) \times D} \}_{i=1}^{M}$ using the original model, then merges them using the estimated object relevance $\{ {\rm A}_{\rm img}^{i} \in \mathbb{R}^{(H \cdot W)} \}_{i=1}^{M}$ as weights:

\begin{equation}
    \mathbf{Z}_{\rm new} = \mathbf{Z}_{\rm text} + \sum\limits_{i=1}^{M} \frac{{\rm A}_{\rm img}^{i}}{\bar {\rm A}_{\rm img}^{i}} \odot \mathbf{Z}_{\rm img}^{i},
    \nonumber
\end{equation}

where $\odot$ is element-wise multiplication with ${\rm A}_{\rm img}^{i}[p,q] \in \mathbb{R}$ and $\mathbf{Z}_{\rm img}^{i}[p,q] \in \mathbb{R}^{D}$~($p \in \{1, 2, \ldots, H\}, q \in \{1, 2, \ldots, W\}$) as each element-pair.
Here, $\bar {\rm A}_{\rm img}^{i} \in \mathbb{R}$ is the average of ${\rm A}_{\rm img}^{i}$, and the division operation is used for normalization~(\textit{i.e.}, the average value of $\frac{{\rm A}_{\rm img}^{i}}{\bar {\rm A}_{\rm img}^{i}}$ equals $1$).
This method adds each $\mathbf{Z}_{\rm img}^{i}$ more to the positions in $\mathbf{Z}_{\rm text}$ with higher relevance to the corresponding object, thus incorporating reference image information more accurately into the corresponding object and mitigating object confusion.
\autoref{tab:training_free_methods} shows that this weighted-merge method can remarkably improve the performance of multi-object personalized image generation on the pre-trained IP-Adapter.

\begin{table}
\renewcommand\arraystretch{1}
\small
\centering
\setlength{\tabcolsep}{1.2mm}{
\begin{tabular}{c *4{c}}
  \toprule
\textbf{\small Merging Method} & \textbf{\small $S_{\rm object\_relevance}$} & \textbf{\small CLIP-T} & \textbf{\small CLIP-I} & \textbf{\small DINO} \\

\midrule
Uniform-Merge & 1.33 & 0.6343 & 0.6409 & 0.3481 \\
Weighted-Merge & \textbf{1.66} & \textbf{0.6427} & \textbf{0.6503} & \textbf{0.3624} \\

\bottomrule
\end{tabular}}
\caption{The performance of different merging methods for the pre-trained IP-Adapter~(training-free) on \textit{Concept101}.}
\vspace{-1em}
\label{tab:training_free_methods}

\end{table}

\subsection{4. Verification with Object Relevance Score \label{sec:verification}}

To verify ${\rm A}_{\rm img}^{i}$ accurately represents the object relevance of each position in $\mathbf{Z}_{\rm text}$, we conduct an experiment in the original text-to-image diffusion model that evaluates the object relevance score $S_{\rm object\_relevance}$ by adding noise to $\mathbf{Z}_{\rm text}$.
Detailedly, we calculate $S_{\rm object\_relevance}$ in three steps: (1) Generate the bounding box ${\rm bbox}_{\mybold{x}}$ of the target object in the generated image $\mybold{x}$ using the Grounding DINO~\cite{liu2023grounding} detection model.
(2) Let $\mybold{x}_{\rm noise}$ denote the generated image with noise added on $\mathbf{Z}_{\rm text}$, $\mybold{x}_{\rm no\_noise}$ denote the generated image without adding noise, then calculate $\Delta_{\mybold{x}}^{\rm bbox}$ as the averaged difference between the pixels of ${\rm bbox}_{\mybold{x}}$ in $\mybold{x}_{\rm noise}$ and $\mybold{x}_{\rm no\_noise}$.
$\Delta_{\mybold{x}}^{\rm non\_bbox}$ is calculated likewise for the region outside the bounding box ${\rm bbox}_{\mybold{x}}$.
(3) Finally, $S_{\rm object\_relevance}$ is calculated as the ratio between the $\Delta_{\mybold{x}}^{\rm bbox}$ and $\Delta_{\mybold{x}}^{\rm non\_bbox}$ averagely over all generated images $\mathcal{X}$~($\| \cdot \|$ denotes cardinality of a set):

\begin{equation}
    S_{\rm object\_relevance} = \frac{1}{\| \mathcal{X} \|} \sum\limits_{\mybold{x} \in \mathcal{X}} \frac{ \Delta_{\mybold{x}}^{\rm bbox} }{ \Delta_{\mybold{x}}^{\rm non\_bbox} }.
    \nonumber
\end{equation}

Therefore, higher $S_{\rm object\_relevance}$ indicates that the added noise has a higher impact on the target object compared to other regions.
We conduct this experiment on the total 1212 text prompts from Concept101 dataset~\cite{kumari2023custom_diffusion}, and the seed for generating each pair of $\mybold{x}_{\rm noise} \in \mathbb{R}^{(H \cdot W) \times D}$ and $\mybold{x}_{\rm no\_noise}$ is set to the same.
Two strategies for adding noise are compared: \textbf{uniform-merge} and \textbf{weighted-merge}.
Uniform-merge directly adds the noise $\epsilon_{\rm object}$ equally into all positions of $\mathbf{Z}_{\rm text}$, while weighted-merge adds the noise with different weights on different positions: $\epsilon_{\rm object} \odot ({\rm A}_{\rm img}^{i} / \bar {\rm A}_{\rm img}^{i})$.
Note that the norm of the sampled noise is equal for these two strategies for fair comparison.
As shown in \autoref{tab:training_free_methods}, $S_{\rm object\_relevance} > 1$ for uniform-merge, indicating that the background of $\mathbf{Z}_{\rm text}$ also has a great influence on the object.
Besides, weighted-merge achieves significantly higher $S_{\rm object\_relevance}$ than uniform-merge, implying that weighted-merge can effectively estimate the object relevance on $\mathbf{Z}_{\rm text}$.

\subsection{5. Training-Based Personalized Image Generation \label{sec:training_framework}}

However, this training-free weighted-merge method still lags behind other multi-object personalized image generation methods, because:
(1) The pre-trained model are trained with only one reference image as input, and directly adding multiple $\mathbf{Z}_{\rm img}^{i}$ will easily disrupt $\mathbf{Z}_{\rm new}$ from its original feature distribution and decrease the quality of the generated images.
(2) Different $\mathbf{Z}_{\rm img}^{i}$ may still conflict at the same position in $\mathbf{Z}_{\rm text}$ when the corresponding values of ${\rm A}_{\rm img}^{i}$ are both high.
To tackle these problems, we propose to continue to train the model with the weighted-merge method on a multi-object dataset, which is to align $\mathbf{Z}_{\rm text} + \sum\limits_{i=1}^{M} \frac{{\rm A}_{\rm img}^{i}}{\bar {\rm A}_{\rm img}^{i}} \odot \mathbf{Z}_{\rm img}^{i}$ with the original feature distribution for higher image quality and alleviate the conflict of different $\mathbf{Z}_{\rm img}^{i}$.

To this end, we first construct the multi-object dataset from the open-sourced SA-1B dataset, following the data-construction paradigm of Subject-Diffusion.
This data-construction paradigm adopts the pre-trained BLIP2~\cite{li2023blip2}, Grounding DINO~\cite{liu2023grounding}, and SAM~\cite{kirillov2023sam} to generate the text prompts, bounding boxes, and segmentation maps of each image.
Furthermore, we propose an object quality score $S_{\rm object\_quality}$ to estimate the object quality of each image and accordingly select the images with high $S_{\rm object\_quality}$.
Detailedly, $S_{\rm object\_quality}$ is calculated based on two factors:
(1) the quality of each individual object;
(2) the quality of each pair of objects.
The first factor is to ensure that the image of each object~(cropped from the original image) is consistent with the object text.
The second factor is to select the object pairs with \textbf{lower similarities}, which facilitates the model to resolve the conflict between multiple reference images and mitigate the object confusion problem, instead of continuing wrongly adding the information of \textbf{another similar} object into the current object.
We utilize the CLIP model $g$ to assess these two factors because of its excellent cross-modal ability.
Let $\mathcal{O}_{\mybold{x}}$ denote the objects in image $\mybold{x}$, $g_{\rm text}(o) \in \mathbb{R}^{D_{\rm clip}}$ and $g_{\rm img}(o) \in \mathbb{R}^{D_{\rm clip}}$ denote the text and image features of object $o$, then $S_{\rm object\_quality}$ of image $\mybold{x}$ is calculated as below~($\cos(\cdot, \cdot)$ denotes cosine similarity):

\begin{equation}
    \nonumber
    \begin{cases}
        S_{\rm object\_quality} = S_{\rm single\_object} + S_{\rm object\_pair}. \\
        S_{\rm single\_object} = \frac{1}{\mathcal{N}_{1}} \sum\limits_{o \in \mathcal{O}_{\mybold{x}}} {\rm cos}(g_{\rm text}(o), g_{\rm img}(o)). \\
        S_{\rm object\_pair} = -\frac{1}{\mathcal{N}_{2}} \sum\limits_{o' \!, o'' \in \mathcal{O}_{\mybold{x}} ; o' \! \ne o''} \cos(g_{\rm img}(o'), g_{\rm img}(o'')).
    \end{cases}
\end{equation}

Here, $\mathcal{N}_{1} = \| \mathcal{O}_{\mybold{x}} \|$ and $\mathcal{N}_{2} = \| \mathcal{O}_{\mybold{x}} \| (\| \mathcal{O}_{\mybold{x}} \| - 1)$ are the normalization terms.
Detailedly, for multi-object personalized image generation, we first filter 215,789 images with multiple annotated objects using the data construction paradigm of Subject-Diffusion, then utilize 100,000 images with the highest $S_{\rm object\_quality}$ for training.

\vspace{0.5em}
\noindent \textbf{Model Architecture.}
\autoref{fig:method} demonstrates the whole pipeline of our method.
We follow previous methods to freeze the original text-to-image diffusion model and only train the parameters~($\mathbf{W}_{\rm img}^{\mathbf{K}}$ and $\mathbf{W}_{\rm img}^{\mathbf{V}}$) for generating each $\mathbf{Z}_{\rm img}^{i}$ in each layer.
Note that $\mathbf{W}_{\rm img}^{\mathbf{K}}$ and $\mathbf{W}_{\rm img}^{\mathbf{V}}$ are shared for generating each $\mathbf{Z}_{\rm img}^{i}$ to save training cost.
Besides, we propose another weighted-merge method to predict the relevance of each position in $\mathbf{Z}_{\rm text}$ to object-unrelated texts, which is to resolve the conflict between $\mathbf{Z}_{\rm text}$ and $\{ \mathbf{Z}_{\rm img}^{i} \}_{i=1}^{M}$.
However, it is difficult to directly extract the text features of these object-unrelated texts and calculate the corresponding cross-attention matrix like ${\rm A}_{\rm img}^{i}$.
To address this problem, this work proposes to predict the weight for the text features with a trainable prediction layer.
Specifically, let $f(\mathbf{Z}_{\rm text}) \in \mathbb{R}^{(H \cdot W)}$ denote the predicted weight for $\mathbf{Z}_{\rm text}$~($f$ is the trainable linear layer followed with a Sigmoid activation function), then $\mathbf{Z}_{\rm new}$ is calculated as below:

\begin{equation}
    \mathbf{Z}_{\rm new} = \frac{ f(\mathbf{Z}_{\rm text}) }{ \bar f(\mathbf{Z}_{\rm text}) } \odot \mathbf{Z}_{\rm text} + \sum\limits_{i=1}^{M} \frac{{\rm A}_{\rm img}^{i}}{\bar {\rm A}_{\rm img}^{i}} \odot \mathbf{Z}_{\rm img}^{i}.
    \nonumber
\end{equation}

\noindent \textbf{Single-Object Personalized Image Generation.}
Our weighted-merge training framework can be extended to other scenarios of simultaneous merging multiple conditions, such as single-object personalized image generation with multiple reference images.
In real applications, a single object may have multiple reference images~(\textit{e.g.}, each object has 4 to 6 reference images in the DreamBooth dataset).
However, previous decoupled cross-attention approaches can only use a single reference image or simply average the features of multiple images, without fully utilizing the information from different reference images.
To tackle this problem, we continue to train the models using our weighted-merge training framework, which enables the model to extract diverse useful information from different reference images and adaptively merge them to achieve superior results.

\section{Experiments}

\begin{table}
\renewcommand\arraystretch{1}
\small
\centering
\setlength{\tabcolsep}{1.2mm}{
\begin{tabular}{c *4{c}}
  \toprule
\textbf{\small Method} & \textbf{\small Type} & \textbf{\small CLIP-T} & \textbf{\small CLIP-I} & \textbf{\small DINO} \\

\midrule
{\small DreamBooth $\bullet$} & FT & 0.7383 & 0.6636 & 0.3849 \\
{\small Custom Diffusion~(Opt) $\bullet$} & FT & 0.7599 & 0.6595 & 0.3684 \\
{\small Custom Diffusion~(Joint)} $\bullet$ & FT & 0.7534 & 0.6704 & 0.3799 \\
{\small Mix-of-Show} $\mybold{\S}$ & FT & 0.7280 & 0.6700 & 0.3940 \\
{\small MC$^2$} $\mybold{\S}$ & FT & 0.7670 & 0.6860 & 0.4060 \\

\midrule
{\small FastComposer} $\mybold{\star}$ & no-FT & 0.7456 & 0.6552 & 0.3574 \\
{\small $\lambda$-ECLIPSE} $\mybold{\star}$ & no-FT & 0.7275 & 0.6902 & 0.3902 \\
{\small ELITE} $\mybold{\star}$ & no-FT & 0.6814 & 0.6460 & 0.3347 \\
{\small IP-Adapter} $\mybold{\star}$ & no-FT & 0.6343 & 0.6409 & 0.3481 \\
{\small SSR-Encoder} $\mybold{\star}$ & no-FT & 0.7363 & 0.6895 & 0.3970 \\

\midrule
{\small \bf Ours~(sdxl)} & no-FT & 0.7750 & 0.6943 & 0.4127 \\
{\small \bf Ours~(sdxl\_plus)} & no-FT & \textbf{0.7765} & \textbf{0.6950} & \textbf{0.4397} \\

\bottomrule
\end{tabular}}
\caption{Performance comparison for multi-object personalized generation on \textit{Concept101}. Here, ``FT'' denotes finetuning-based method, ``no-FT'' denotes finetuning-free method, and bold font denotes the best result. Each CLIP-T score is multiplied by 2.5 following Custom Diffusion.}
\label{tab:benchmark_concept101_multi}

\end{table}
\begin{table}
\renewcommand\arraystretch{1}
\small
\centering
\setlength{\tabcolsep}{2.2mm}{
\begin{tabular}{c *4{c}}
  \toprule
\textbf{\small Method} & \textbf{\small Type} & \textbf{\small CLIP-T} & \textbf{\small CLIP-I} & \textbf{\small DINO} \\

\midrule
{\small DreamBooth} $\dagger$ & FT & 0.308 & 0.695 & 0.430 \\
{\small Custom Diffusion} $\dagger$ & FT & 0.300 & 0.698 & 0.464 \\
{\small Subject Diffusion} $\dagger$ & no-FT & 0.310 & 0.696 & \textbf{0.506} \\

\midrule
{\small \bf Ours~(sdxl)} & no-FT & \textbf{0.311} & \textbf{0.726} & 0.482 \\

\bottomrule
\end{tabular}}
\caption{Performance comparison for multi-object personalized generation on \textit{DreamBooth}.}
\vspace{-1em}
\label{tab:benchmark_dreambooth_multi}

\end{table}


\vspace{0.5em}
\noindent \textbf{Implementation details.}
Our main experiments are conducted on the pre-trained IP-Adapter with sdxl model~\cite{podell2023sdxl} and sdxl\_plus model~\cite{jaegle2021perceiver} as the text-to-image diffusion models and OpenCLIP ViT-bigG/14 as the image encoder.
The parameters of sdxl \& sdxl\_plus model and image encoder are frozen, and only the parameters for projecting image features and predicting text weights are trainable.
During training, we adopt AdamW optimizer with a learning rate of 1e-4, and train the model on 8 PPUs for 30,000 steps with a batch size of 4 per PPU.
To enable classifier-free guidance, we use a probability of 0.05 to drop text and image individually, and a probability of 0.05 to drop text and image simultaneously.
During inference, we adopt DDIM sampler with 50 steps and set the guidance scale to 7.5.
We also conduct experiments on other pre-trained models based on decoupled cross-attention to verify the generalization ability of our method, in \textbf{S2.2} of the appendix.

\vspace{0.5em}
\noindent \textbf{Test benchmark.}
For multi-object personalized image generation, we follow the \textit{Concept101}~\cite{kumari2023custom_diffusion} benchmark that has evaluated many methods.
Besides, we also evaluate our method on the \textit{DreamBooth} benchmark for comparison with Subject-Diffusion.

\vspace{0.5em}
\noindent \textbf{Evaluation metrics.}
We follow previous methods to adopt three metrics~(CLIP-T, CLIP-I, and DINO) for evaluation.
Specifically, CLIP-T evaluates the similarity between the generated images and given text prompts;
CLIP-I and DINO evaluate the similarity between the generated images and the reference images.
5 images are generated for each prompt to ensure the evaluation stability.

\vspace{0.5em}
\noindent \textbf{Baseline methods.}
We compare our method with both finetuning-based methods~(\textit{e.g.}, Textual Inversion, DreamBooth, Custom Diffusion, MC$^2$) and finetuning-free methods~(\textit{e.g.}, SSR-Encoder, Subject-Diffusion).

\begin{table}
\renewcommand\arraystretch{1}
\small
\centering
\setlength{\tabcolsep}{2.9mm}{
\begin{tabular}{c *3{c}}
  \toprule
\textbf{\small Method} & \textbf{\small CLIP-T} & \textbf{\small CLIP-I} & \textbf{\small DINO} \\

\midrule
{\small Uniformly Add} & 0.7702 & 0.6816 & 0.3937 \\
{\small Locally Add} & 0.7732 & 0.6851 & 0.3958 \\
{\small + Image Weight} & 0.7734 & 0.6940 & 0.4079 \\
{\small + Text Weight} & 0.7726 & 0.6924 & 0.4032 \\
{\small + Image \& Text Weights} & \textbf{0.7750} & \textbf{0.6943} & \textbf{0.4127} \\

\bottomrule
\end{tabular}}
\caption{Ablation experiments of weighted-merge methods for multi-object personalized generation on \textit{Concept101}.}
\label{tab:ablation_weighted_method}

\end{table}
\begin{table}
\renewcommand\arraystretch{1}
\small
\centering
\setlength{\tabcolsep}{0.5mm}{
\begin{tabular}{c *3{c}}
  \toprule
\textbf{\small Method} & \textbf{\small CLIP-T} & \textbf{\small CLIP-I} & \textbf{\small DINO} \\

\midrule
{\small 100,000 images~(lowest $S_{\rm object\_pair}$)} & 0.7708 & 0.6880 & 0.3963 \\
{\small 100,000 images~(highest $S_{\rm object\_pair}$)} & 0.7733 & 0.6923 & 0.4056 \\
{\small 100,000 images~(highest $S_{\rm object\_quality}$)} & \textbf{0.7750} & \textbf{0.6943} & \textbf{0.4127} \\

\bottomrule
\end{tabular}}
\caption{Ablation experiments of image selection strategies for multi-object personalized generation on \textit{Concept101}.}
\vspace{-1em}
\label{tab:ablation_image_selection}

\end{table}

\begin{figure*}[t]
\centering
    \includegraphics[width=\linewidth]{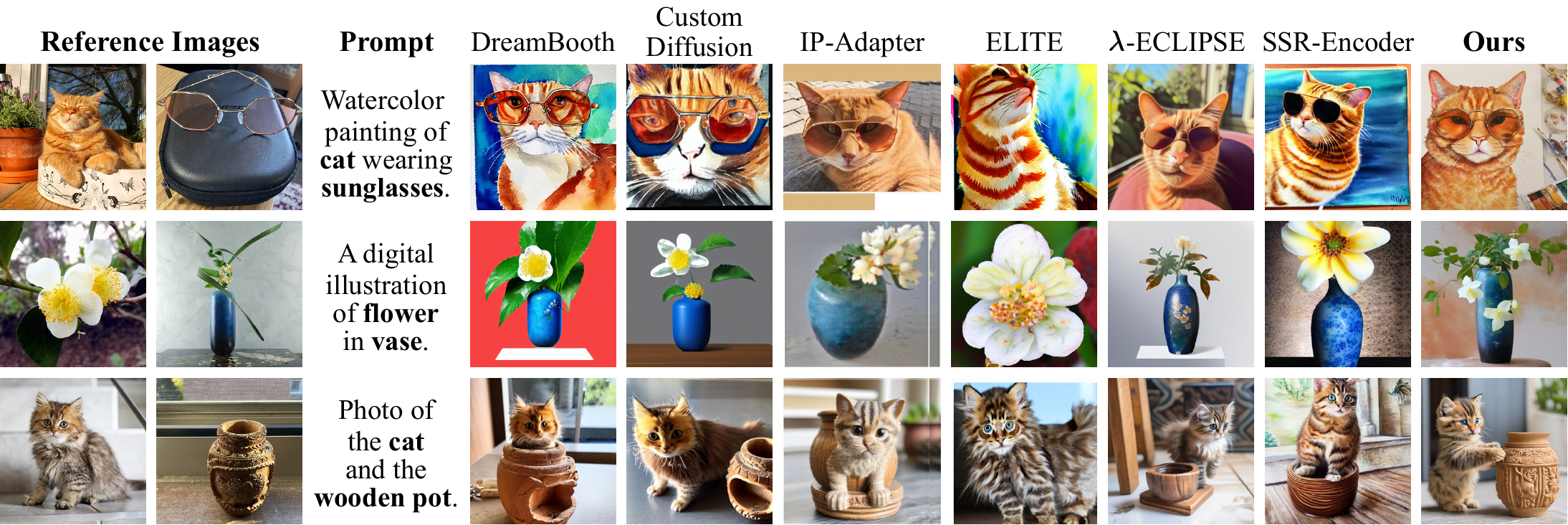}
\caption{Qualitative comparisons of different methods on multi-object personalized image generation.}
\vspace{-1.5em}
\label{fig:visualization}
\end{figure*}
\begin{figure}[t]
\centering
    \includegraphics[width=\linewidth]{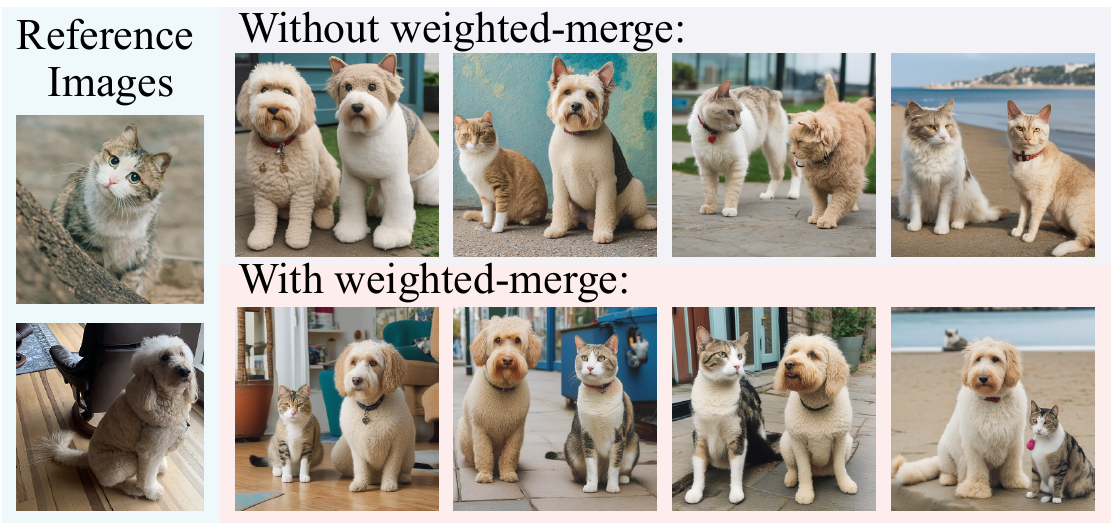}
\caption{Qualitative ablation experiment.}
\vspace{-1.5em}
\label{fig:ablation_weighted_merge}
\end{figure}

\subsection{Multi-Object Personalized Generation}

We conduct both quantitative and qualitative comparisons between our method and baseline methods.

\vspace{0.5em}
\noindent \textbf{Quantitative Comparisons.}
\autoref{tab:benchmark_concept101_multi} demonstrates the quantitative results of different methods on \textit{Concept101}.
Note that the results of methods marked with $\bullet$ are from the GitHub page of Custom Diffusion~\cite{kumari2023custom_diffusion}, the results of methods marked with $\mybold{\S}$ are from the paper of MC$^2$~\cite{jiang2024mc}, and the results of methods marked with $\mybold{\star}$ are re-implemented faithfully following their released code and weights~(their original evaluation datasets have not been made public).

As shown in \autoref{tab:benchmark_concept101_multi}, early finetuning-free methods~(\textit{e.g.}, FastComposer, $\lambda$-ECLIPSE) achieve inferior performance because they merely incorporate the image features into the text embeddings, without fully utilizing the image information.
Recent methods enhance the utilization of image information with decoupled cross-attention to integrate image features into the middle layers of the model, but they have yet to achieve satisfactory results due to the object confusion problem.
Differently, our method generalizes decoupled cross-attention to merging multiple reference images by resolving the object confusion problem, which achieves significantly superior performance to existing methods.

\autoref{tab:benchmark_dreambooth_multi} demonstrates the quantitative results of different methods on the DreamBooth dataset. The results of methods marked with $\dagger$ are from the paper of Subject-Diffusion.
In this benchmark, our method outperforms Subject-Diffusion in 2 of 3 evaluation metrics, and surpasses it in the CLIP-I score by a large margin~(0.726 \textit{vs.} 0.696).

\vspace{0.5em}
\noindent \textbf{Qualitative Comparisons.}
\autoref{fig:visualization} demonstrates the qualitative results of different methods on \textit{Concept101}.
The results of the original IP-Adapter indicate that it generates images with low image quality, due to the object confusion problem and the distortion of feature distribution when merging multiple images once.
Next, after employing the weighted-merge training framework on the original IP-Adapter, our method can generate images with high image quality and mitigate object confusion, realizing the best qualitative results.

Besides, we provide more visualization results of our method in \textbf{S3} of the appendix~(\textit{e.g.}, simultaneously merging \textbf{more than} two objects).

\subsection{Single-Object Personalized Generation}

\begin{figure}[t]
\centering
    \includegraphics[width=\linewidth]{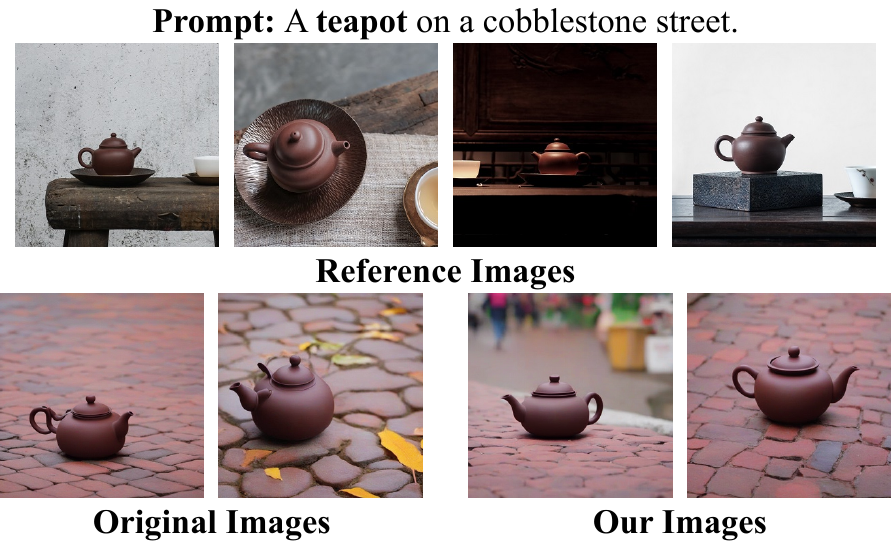}
\caption{An example of visualizations of single-object personalized image generation with multiple reference images.}
\vspace{-1em}
\label{fig:single_object_intro}
\end{figure}

For single-object personalized image generation, we utilize the proposed $S_{\rm single\_object}$~($S_{\rm object\_pair}$ is eliminated in the single-object scenario) to select 100,000 high-quality images for training.
As shown in \autoref{tab:benchmark_dreambooth_single}, our weighted-merge framework can improve all three scores of the original IP-Adapter and ELITE on the DreamBooth dataset.
Besides, \autoref{fig:single_object_intro} shows the qualitative comparisons between our model and the original model, implying that our model can capture important image information from different images, instead of ignoring the unique details of some images by the original model.

\subsection{Ablation Experiments}

\noindent \textbf{Weighted-Merge Training Framework.}
We conduct ablation experiments on two proposed weight estimation methods~(text weight $f(\mathbf{Z}_{\rm text})$ \& image weight $\{ {\rm A}_{\rm img}^{i} \}_{i=1}^{M}$) of the weighted-merge training framework with sdxl model as the backbone.
\autoref{tab:ablation_weighted_method} demonstrates that locally adding reference image features does not show obvious improvement compared to uniform adding.
Besides, \autoref{tab:ablation_weighted_method} indicates that these two weight estimation methods effectively enhance the performance of multi-object personalized generation, and the best performance is achieved when they are simultaneously used. 
Moreover, the qualitative ablation experiment in \autoref{fig:ablation_weighted_merge} also verifies the effectiveness of our weighted-merge method with the visualization results.
Detailedly, the images generated without weighted-merge blend the reference image features of different objects, while the images generated with weighted-merge can accurately map the reference image features to their corresponding objects.

\noindent \textbf{Image Selection.}
\autoref{tab:ablation_image_selection} shows the performance of multi-object personalized generation with different image selection strategies~(with sdxl model as the backbone), implying that the images selected by our proposed $S_{\rm object\_quality}$ lead to superior results.

\begin{figure}[t]
\centering
    \includegraphics[width=\linewidth]{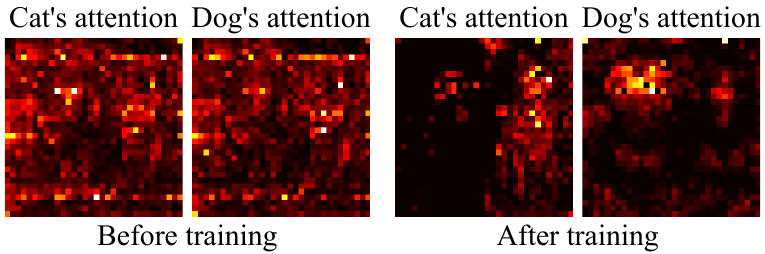}
\caption{The attention maps of reference image features on the latent image features $\mathbf{Z}$ before/after training.}
\vspace{-1.5em}
\label{fig:ablation_attn_map}
\end{figure}

\noindent \textbf{Change of Attention Maps.}
We calculate the attention maps between reference image features of two objects~(cat \& dog from \autoref{fig:ablation_weighted_merge}) and the latent image features $\mathbf{Z}$ in the middle cross-attention layer.
As shown in \autoref{fig:ablation_attn_map}, the attention maps of the two objects become more distinct after training, thereby alleviating the object confusion problem.

Furthermore, we provide ablation experiments~(\textit{e.g.}, the number of training images) in \textbf{S2.3} of the appendix.

\begin{table}[t]
\renewcommand\arraystretch{1}
\small
\centering
\setlength{\tabcolsep}{1.7mm}{
\begin{tabular}{c *4{c}}
  \toprule
\textbf{\small Method} & \textbf{\small Type} & \textbf{\small CLIP-T} & \textbf{\small CLIP-I} & \textbf{\small DINO} \\

\midrule

{\small Textual Inversion~$\dagger$} & FT & 0.255 & 0.780 & 0.569 \\
{\small DreamBooth~$\dagger$} & FT & 0.305 & 0.803 & 0.668 \\
{\small Break-A-Scene~$\dagger$} & FT & 0.287 & 0.788 & 0.653 \\
{\small BLIP-Diffusion~$\dagger$} & no-FT & 0.300 & 0.779 & 0.594 \\

\midrule
{\small IP-Adapter~(Original)~$\dagger$} & no-FT & 0.274 & 0.809 & 0.608 \\
{\small IP-Adapter~(\textbf{Ours})} & no-FT & \textbf{0.296} & \textbf{0.812} & \textbf{0.620} \\

\midrule
{\small ELITE~(Original)~$\dagger$} & no-FT & 0.298 & 0.775 & 0.605 \\
{\small ELITE~(\textbf{Ours})} & no-FT & \textbf{0.304} & \textbf{0.788} & \textbf{0.622} \\

\bottomrule
\end{tabular}}
\caption{Performance comparison for single-object personalized generation on \textit{DreamBooth}. Here, ``FT'' denotes finetuning-based method, ``no-FT'' denotes finetuning-free method, and bold font denotes the best result compared to the original finetuning-free method.}
\label{tab:benchmark_dreambooth_single}

\end{table}

\section{Conclusion}

In this work, we generalize the finetuning-free methods with decoupled cross-attention for merging multiple reference images, by mitigating the object confusion problem.
To this end, we explore the importance of various positions of latent image features in relation to the target object within the diffusion model, and accordingly propose a weighted-merge method to integrate reference image features with their corresponding objects.
This weighted-merge method can directly improve the performance on multi-object generation of existing pre-trained models in a training-free manner.
Next, we continue to train the pre-trained models on a multi-object dataset constructed with a proposed object quality score to further enhance the performance.
Besides, our weighted-merge training framework can be applied to single-object generation when a single object has multiple reference images.
Experiment results demonstrate that our method achieves significantly superior performance to existing methods.
We hope our method and dataset~(will be made publicly available) can contribute to the community of personalized image generation.

{\small
\bibliographystyle{ieee_fullname}
\bibliography{11_references}
}


\end{document}